\documentclass[journal]{IEEEtran}

\ifCLASSINFOpdf
\else
   \usepackage[dvips]{graphicx}
\fi
\usepackage{url}

\usepackage{graphicx}
\usepackage{bm}
\usepackage{amsmath}
\usepackage{booktabs}
\usepackage{multirow}
\usepackage{amsfonts}

\begin{document}

\title{TBFormer: Two-Branch Transformer for Image Forgery Localization}

\author{{Yaqi Liu, Binbin Lv, Xin Jin, Xiaoyu Chen, and Xiaokun Zhang}
\thanks{Manuscript received February 25, 2023. This work was supported in part by the NSFC under Grant 62102010; in part by the Advanced Discipline Construction Project of Beijing Universities under Grant 20210037Z0401; and in part by the Information Center Project of China North Industries Group Corporation Limited under Grant 20220100H0113. {\em (Corresponding author: Xin Jin.)}}
\thanks{Yaqi Liu, Binbin Lv, Xin Jin, and Xiaokun Zhang are with the Beijing Electronic Science and Technology Institute, Beijing 100070, China (e-mail: liuyaqi@besti.edu.cn; lv-bin-bin@outlook.com; jinxin@besti.edu.cn; sam@besti.edu.cn).}
\thanks{Xiaoyu Chen is with the China North Industries Group Corporation Limited, Beijing 100089, China.}}

\markboth{Journal of \LaTeX\ Class Files, Vol. 14, No. 8, August 2015}
{Shell \MakeLowercase{\textit{et al.}}: Bare Demo of IEEEtran.cls for IEEE Journals}
\maketitle

\begin{abstract}
Image forgery localization aims to identify forged regions by capturing subtle traces from high-quality discriminative features. In this paper, we propose a Transformer-style network with two feature extraction branches for image forgery localization, and it is named as Two-Branch Transformer (TBFormer). Firstly, two feature extraction branches are elaborately designed, taking advantage of the discriminative stacked Transformer layers, for both RGB and noise domain features. Secondly, an Attention-aware Hierarchical-feature Fusion Module (AHFM) is proposed to effectively fuse hierarchical features from two different domains. Although the two feature extraction branches have the same architecture, their features have significant differences since they are extracted from different domains. We adopt position attention to embed them into a unified feature domain for hierarchical feature investigation. Finally, a Transformer decoder is constructed for feature reconstruction to generate the predicted mask. Extensive experiments on publicly available datasets demonstrate the effectiveness of the proposed model.
\end{abstract}


\IEEEpeerreviewmaketitle

\section{Introduction}

\IEEEPARstart{E}{diting} digital images may change the semantic content of original images, and the edited images are often too realistic to distinguish their authenticity. It poses a threat to the stability and harmony of the society if they are used illegally. Image forgery localization is a kind of image forensics task which aims at locating forged regions in investigated images, and it has attracted more and more attention in both research and industry \cite{ref51}-\cite{ref52}.

Researchers have proposed many image forgery localization methods for specific forgery types, e.g., splicing \cite{ref1}-\cite{ref5}, copy-move \cite{ref7}-\cite{ref12}, and removal \cite{ref13}-\cite{ref14}. In practice, the investigated image may contain multiple forgery types at the same time \cite{ref15}-\cite{ref16}. Some researchers \cite{ref15}-\cite{ref20} have also proposed methods applicable to multiple forgery types, while many of these methods extract features from the RGB domain \cite{ref16}. Some researchers \cite{ref14}-\cite{ref15}, \cite{ref21}-\cite{ref26} have also attempted to combine features extracted from different domains. Wu et al. \cite{ref25} and Hu et al. \cite{ref15} concated RGB image and its corresponding noise map before the feature extractor. Zhou et al. \cite{ref26} and Chen et al. \cite{ref21} designed two parallel branches to extract RGB features and noise features. While the above-mentioned methods are constructed based on convolutional neural networks.

In recent years, Transformer has been widely used in various vision tasks, e.g., object detection \cite{ref27}-\cite{ref30} and image segmentation \cite{ref31}-\cite{ref33}, showing superior performance. Researchers also tried to apply Transformer to image forgery localization. Wang et al. \cite{ref35} designed a multimodal Transformer framework. Instead of using images directly as the input, they used convolutional layers to extract feature maps for patch embedding. Sun et al. \cite{ref38} adopted multiple Transformer layers to extract features only from the RGB domain, and constructed a convolutional decoder.

\begin{figure*}
\centerline{\includegraphics[width=0.85\textwidth]{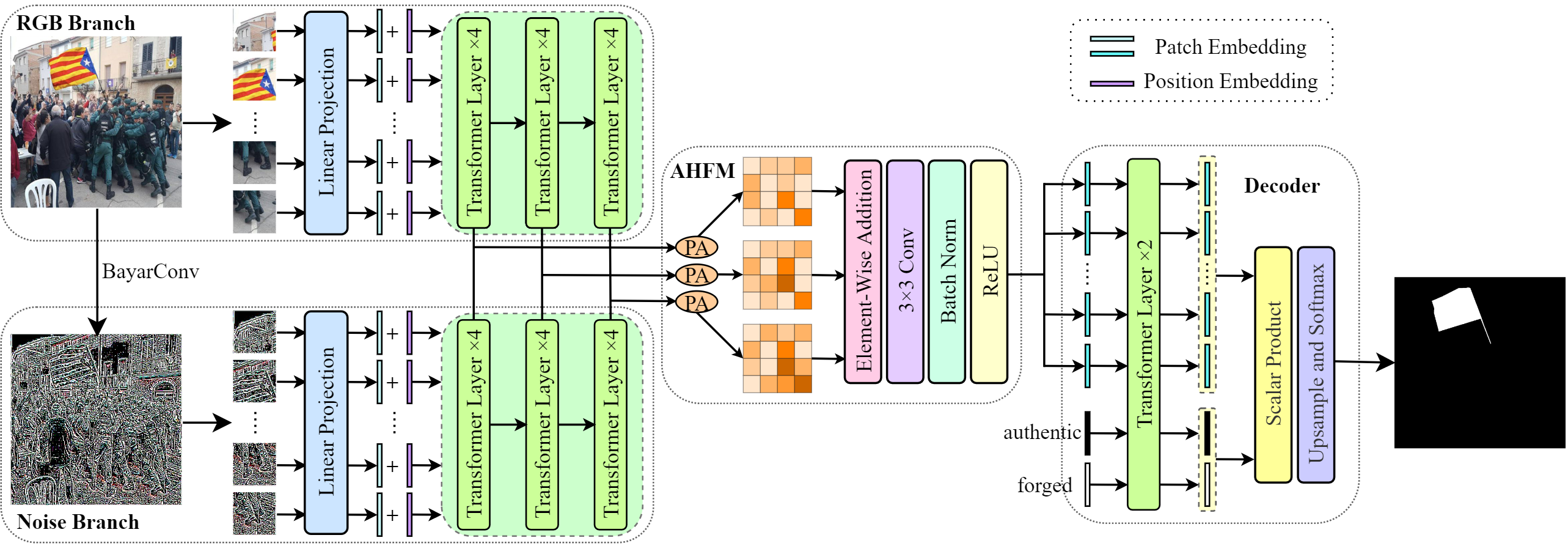}}
\vspace{-1.0em}
\caption{Overview of the proposed TBFormer. TBFormer consists of two feature extraction branches, an AHFM module, and a Transformer decoder.}
\label{fig1}
\vspace{-1.0em}
\end{figure*}

In this paper, we propose a Transformer-style image forgery localization network, namely TBFormer, and the architecture is shown in Fig. \ref{fig1}. The noise domain contains subtle forgery traces, which are visually invisible and difficult to capture from the RGB domain. Therefore, we develop two feature extraction branches with multiple Transformer layers to extract discriminative features from the RGB domain and the noise domain independently. The two branches have the same architecture and their weights are not shared. The consideration is that Transformer layers are powerful for discriminative feature representation, and the non-shared design makes them concentrate on their specific domains. However, the non-shared feature extraction branches provide feature maps from different domains with large differences. How to fuse these feature maps becomes a key problem. Thus, we design an Attention-aware Hierarchical-feature Fusion Module (AHFM) to effectively fuse hierarchical features from two different domains. RGB features and noise features from the same layer are gone through a position attention module, to integrate them into a unified feature domain. Then hierarchical features are combined by element-wise addition and a convolutional layer to get the final fused feature map with rich hierarchical information from both RGB and noise domains. Finally, we design a Transformer decoder to reconstruct the fused features and provide the predicted mask. Category embeddings are set in the decoder to further learn unified feature representations of authentic and forged classes, and they are interacted with fused feature map patch embeddings to produce the predicted masks. Last but not least, in order to train and test our Transformer-style network, we generate a synthesized image dataset with 140432 images for training, 7787 images for validating, and 7787 images for testing. The synthesized dataset is made publicly available for further research.

The main contributions of this paper can be summarized as follows: (1) A novel Transformer-style network (TBFormer) with two feature extraction branches is proposed for image forgery localization. (2) An Attention-aware Hierarchical-feature Fusion Module (AHFM) is proposed to effectively fuse hierarchical features from two different domains. (3) A Transformer decoder is constructed for feature reconstruction to generate the predicted mask. (4) All our codes, models and the generated dataset are available online (\url{https://github.com/free1dom1/TBFormer}).

\section{PROPOSED METHOD}

\subsection{Two-Branch Feature Extractor}

To exploit the potential forgery cues in different domains, we design two feature extraction branches to extract discriminative features from the RGB domain and the noise domain. The two branches have the same architecture, and their weights are not shared which makes them concentrate on their specific domains. We adopt BayarConv \cite{ref39} for converting the RGB domain to the noise domain. Transformer can overcome the shortcomings of convolutional neural networks with only limited receptive fields and has the powerful ability to model contextual global dependencies \cite{ref36}-\cite{ref37}. The rich contextual information is also crucial for locating forged regions, so Transformer is adopted for our feature extraction.

The input color RGB image ${\bm{I}_{c}\in \mathbb{R}^{H\times W\times 3}}$ is first converted to the noise map ${\bm{I}_{n}\in \mathbb{R}^{H\times W\times 3}}$ by BayarConv, where $W$ and $H$ denote the width and height of the input image. We divide ${\bm{I}_{c}}$ into image patches of size ${16\times 16}$ to obtain the sequence ${\bm{X}_{c}=\left\{\bm{x}_{c}^{\left(1\right)},\bm{x}_{c}^{\left(2\right)},\cdots,\bm{x}_{c}^{\left(N\right)}\right\}}$, where ${\bm{x}_{c}^{\left(i\right)}\in \mathbb{R}^{16\times 16\times 3}}$ and ${N=H/16\times W/16}$ is the number of image patches. Each image patch ${\bm{x}_{c}^{\left(i\right)}}$ is reshaped into a one-dimensional vector, followed by a linear projection layer to obtain the image patch embedding sequence ${\bm{P}_{c}=\left\{\bm{p}_{c}^{\left(1\right)},\bm{p}_{c}^{\left(2\right)},\cdots,\bm{p}_{c}^{\left(N\right)}\right\}\in \mathbb{R}^{N\times L}}$, where $L$ denotes the feature dimension. The corresponding position embedding ${\mathbf{pos}_{c}^{\left(i\right)}}$ is added to the image patch embedding ${\bm{p}_{c}^{\left(i\right)}}$ to obtain the resulting input sequence ${\bm{E}_{c}=\left\{\bm{e}_{c}^{\left(1\right)},\bm{e}_{c}^{\left(2\right)},\cdots,\bm{e}_{c}^{\left(N\right)}\right\}\in \mathbb{R}^{N\times L}}$, where ${\bm{e}_{c}^{\left(i\right)}=\bm{p}_{c}^{\left(i\right)}+\mathbf{pos}_{c}^{\left(i\right)}}$. Then ${\bm{E}_{c}}$ is fed into the feature extractor which is constructed based on $12$ Transformer layers. The feature maps of the $4$th, $8$th, and $12$th layers (i.e., $\bm{T}_{c}^{\left(4\right)},\bm{T}_{c}^{\left(8\right)},\bm{T}_{c}^{\left(12\right)}$) are output for further investigation:
\begin{equation}
\bm{T}_{c}=\left\{\bm{T}_{c}^{\left(4\right)},\bm{T}_{c}^{\left(8\right)},\bm{T}_{c}^{\left(12\right)}\right\}=f_{c}\left(\bm{{E}_{c}}\right)
\end{equation}
where ${f_{c}}$ denotes the feature extractor of the RGB branch. The Transformer layer consists of a Multi-Head Self-Attention (MSA) block and a Multi-Layer Perceptron (MLP) block, and the architecture of the $i$th layer can be represented as: 
\begin{equation}
\bm{M}_{c}^{\left(i\right)}={\rm{MSA}}_{c}^{\left(i\right)}\left({\rm{LN}}\left(\bm{T}_{c}^{\left(i-1\right)}\right)\right)+\bm{T}_{c}^{\left(i-1\right)}
\end{equation}
\begin{equation}
\bm{T}_{c}^{\left(i\right)}={\rm{MLP}}_{c}^{\left(i\right)}\left({\rm{LN}}\left(\bm{M}_{c}^{\left(i\right)}\right)\right)+\bm{M}_{c}^{\left(i\right)}
\end{equation}
where ${\rm{LN}}$ represents layer norm. The ${{\rm{MSA}}_{c}^{\left(i\right)}}$ block is constituted by the Self-Attention (SA) operation:
\begin{equation}
{\rm{SA}}_{c}^{\left(i\right)}\left(\bm{T}_{c}^{\left(i-1\right)}\right)={\rm{softmax}}\left({\bm{Q}}_{c}^{\left(i\right)}\left({\bm{K}}_{c}^{\left(i\right)}\right)^{\rm{T}}/\sqrt{L}\right){\bm{V}}_{c}^{\left(i\right)}
\end{equation}
where query, key, value are computed as ${{\bm{Q}}_{c}^{\left(i\right)}=\bm{T}_{c}^{\left(i-1\right)}{\bm{W}}_{\rm{cQ}}^{\left(i\right)}}$, ${{\bm{K}}_{c}^{\left(i\right)}=\bm{T}_{c}^{\left(i-1\right)}{\bm{W}}_{\rm{cK}}^{\left(i\right)}}$, ${{\bm{V}}_{c}^{\left(i\right)}=\bm{T}_{c}^{\left(i-1\right)}{\bm{W}}_{\rm{cV}}^{\left(i\right)}}$, and ${{\bm{W}}_{\rm{cQ}}^{\left(i\right)}}$, ${{\bm{W}}_{\rm{cK}}^{\left(i\right)}}$, ${{\bm{W}}_{\rm{cV}}^{\left(i\right)}}$ are the learnable parameters of three linear projection layers in self-attention \cite{ref53}.

The same processes are performed on the noise map ${\bm{I}_{n}}$ to obtain ${\bm{E}_{n}\in \mathbb{R}^{N\times L}}$. The noise features are obtained by feeding ${\bm{E}_{n}}$ into the feature extractor of the noise branch:
\begin{equation}
\bm{T}_{n}=\left\{\bm{T}_{n}^{\left(4\right)},\bm{T}_{n}^{\left(8\right)},\bm{T}_{n}^{\left(12\right)}\right\}=f_{n}\left(\bm{{E}_{n}}\right)
\end{equation}
where $f_{n}$ denotes the feature extractor of the noise branch, ${\bm{T}_{n}^{\left(4\right)},\bm{T}_{n}^{\left(8\right)},\bm{T}_{n}^{\left(12\right)}\in \mathbb{R}^{N\times L}}$ denote the features output by the $4$th, $8$th, and $12$th Transformer layers.

\subsection{Attention-aware Hierarchical-feature Fusion Module}

The feature maps of the two branches have significant differences for that they are extracted from different domains. A carefully designed decoder is helpful for mask reconstruction from different domains, and a well-designed feature fusion module is also an indispensable part of a network to investigate multi-domain information. We design an Attention-aware Hierarchical-feature Fusion Module (AHFM) to effectively fuse hierarchical features from two different domains.

\begin{figure}
\centerline{\includegraphics[width=\columnwidth]{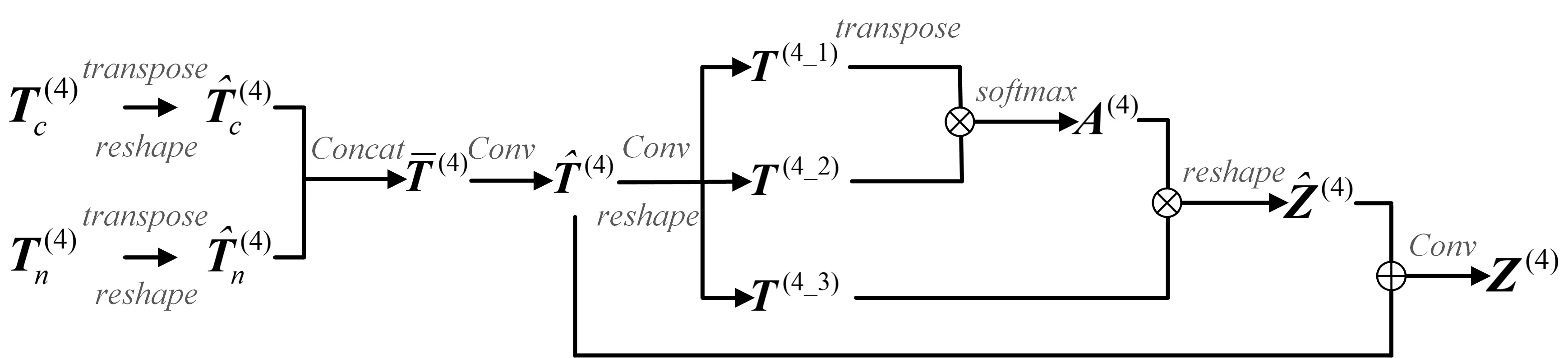}}
\vspace{-1.0em}
\caption{\centering{Computational procedure of Position Attention.}}
\label{fig2}
\vspace{-1.0em}
\end{figure}

For the RGB features and noise features from the same layer, we construct a position attention block \cite{ref41} to investigate their correlation and fuse them into unified feature maps. Taking the $4$th-layer features as examples, the matrixes ${\bm{T}_{c}^{\left(4\right)}\in \mathbb{R}^{N\times L}}$ and ${\bm{T}_{n}^{\left(4\right)}\in \mathbb{R}^{N\times L}}$ are transposed and reshaped to get the three-dimensional tensors ${\hat{\bm{T}}_{c}^{\left(4\right)}\in \mathbb{R}^{L\times h\times w}}$ and ${\hat{\bm{T}}_{n}^{\left(4\right)}\in \mathbb{R}^{L\times h\times w}}$, where ${N=h\times w}$, ${h=H/16}$, and ${w=W/16}$. Then, ${\hat{\bm{T}}_{c}^{\left(4\right)}}$ and ${\hat{\bm{T}}_{n}^{\left(4\right)}}$ are concatenated along the channel dimension to get ${\bar{\bm{T}}^{\left(4\right)}\in \mathbb{R}^{2L\times h\times w}}$. A convolution operation is performed on ${\bar{\bm{T}}^{\left(4\right)}}$ to get ${\hat{\bm{T}}^{\left(4\right)}\in \mathbb{R}^{L\times h\times w}}$, then three different convolutional layers are constructed for ${\hat{\bm{T}}^{\left(4\right)}}$ to obtain ${\hat{\bm{T}}^{\left(4\_1\right)}\in \mathbb{R}^{L/8\times h\times w}}$, ${\hat{\bm{T}}^{\left(4\_2\right)}\in \mathbb{R}^{L/8\times h\times w}}$, and ${\hat{\bm{T}}^{\left(4\_3\right)}\in \mathbb{R}^{L\times h\times w}}$. Then, they are reshaped to ${\bm{T}^{\left(4\_1\right)}\in \mathbb{R}^{L/8\times N}}$, ${\bm{T}^{\left(4\_2\right)}\in \mathbb{R}^{L/8\times N}}$, and ${\bm{T}^{\left(4\_3\right)}\in \mathbb{R}^{L\times N}}$. Position attention weights ${{\bm{A}}^{\left(4\right)}\in \mathbb{R}^{N\times N}}$ can be computed as:
\begin{equation}
{\bm{A}}^{\left(4\right)}={\rm{softmax}}\left(\left(\bm{T}^{\left(4\_1\right)}\right)^{\rm{T}}\bm{T}^{\left(4\_2\right)}\right)
\end{equation}
Then, we conduct matrix multiplication between ${\bm{T}^{\left(4\_3\right)}}$ and ${{\bm{A}}^{\left(4\right)}}$, and the computed result is reshaped to ${\hat{\bm{Z}}^{\left(4\right)}\in \mathbb{R}^{L\times h\times w}}$. Then, $\hat{\bm{Z}}^{\left(4\right)}$ is multiplied with a learnable weight ${{\alpha}^{\left(4\right)}}$, and we perform element-wise addition between the weighted $\hat{\bm{Z}}^{\left(4\right)}$ and ${\hat{\bm{T}}^{\left(4\right)}}$. A convolution operation is conducted to get the fused feature map ${\bm{Z}^{\left(4\right)}}\in \mathbb{R}^{L\times h\times w}$ as follows:
\begin{equation}
\bm{Z}^{\left(4\right)}={\rm{Conv}}^{\left(4\right)}\left({\alpha}^{\left(4\right)} \left(\bm{T}^{\left(4\_3\right)}{\bm{A}}^{\left(4\right)}\right)_{\rm{reshape}}\oplus \hat{\bm{T}}^{\left(4\right)}\right)
\end{equation}
where ${\oplus}$ denotes element-wise addition. The detailed computational procedure is shown in Fig. \ref{fig2}. Following the same computational procedure, we can also get the fused feature maps ${\bm{Z}^{\left(8\right)}}$ for the $8$th layer and ${\bm{Z}^{\left(12\right)}}$ for the $12$th layer. 

In order to sufficiently integrate the hierarchical features, we conduct element-wise addition followed by a convolution operation to get the final fused feature map ${\bm{Z}\in \mathbb{R}^{L\times h\times w}}$:
\begin{equation}
\bm{Z}={\rm{Conv}}\left(\bm{Z}^{\left(12\right)}\oplus \bm{Z}^{\left(8\right)}\oplus \bm{Z}^{\left(4\right)}\right)
\end{equation}
The general framework of our AHFM module is shown in Fig. \ref{fig1} (the bounding box of AHFM).

\subsection{Transformer Decoder}

Image forgery localization classifies each pixel in an image into two classes, i.e., authentic class and forged class. It can essentially be considered as a special image segmentation task. We set two learnable category embeddings in the decoder to further learn the feature representations of authentic and forged classes \cite{ref42}, and they are interacted with the patch embeddings of the fused feature map to produce the predicted masks. Our decoder mainly contains $2$ Transformer layers.

Specifically, ${\bm{Z}\in \mathbb{R}^{L\times h\times w}}$ is sequentially reshaped, transposed, and linearly projected to obtain the embedding sequence ${\dot{\bm{Z}}\in \mathbb{R}^{N\times L}}$. Then $\dot{\bm{Z}}$ and the category embeddings ${\bm{S}\in \mathbb{R}^{2\times L}}$ are reconstructed by Transformer layers to obtain ${\ddot{\bm{Z}}\in \mathbb{R}^{N\times L}}$ and ${\ddot{\bm{S}}\in \mathbb{R}^{2\times L}}$. After performing linear projection and L2 normalization on ${\ddot{\bm{Z}}}$ and ${\ddot{\bm{S}}}$, respectively, the quantization value ${\ddot{\bm{Y}}\in \mathbb{R}^{N\times 2}}$ can be obtained by the scalar product operation:
\begin{equation}
\ddot{\bm{Y}}=L_{2}\left({f_{\rm{proj}}}\left(\ddot{\bm{Z}}\right)\right)\left(L_{2}\left({f_{\rm{proj}}}\left(\ddot{\bm{S}}\right)\right)\right)^T
\end{equation}
where $L_2$ denotes L2 normalization and $f_{\rm{proj}}$ denotes linear projection.
The transpose and reshape operations are performed sequentially on ${\ddot{\bm{Y}}}$ to obtain ${\bm{Y}\in \mathbb{R}^{2\times h\times w}}$, and the predicted mask ${\bm{M}}$ is computed as:
\begin{equation}
\bm{M}={\rm{softmax}}\left({\rm{Upsample}}\left(\bm{Y}\right)\right)
\end{equation}
where $\rm{Upsample}$ denotes the upsampling operation which can resize $\bm{Y}$ to the same size as the input image. Our model is trained using a pixel-level binary cross-entropy loss function.

\section{EXPERIMENTS}

\subsection{Experimental Settings}

\subsubsection{Synthesized dataset}
{We generate a large amount of synthesized images to train our Transformer-style network. For splicing and copy-move operations, we enlarge the CASIA v2.0 dataset \cite{ref43}-\cite{ref44}. By learning the association between scenes and forged regions, we try to find the most concealed position for inserting forged regions. Specifically, we select the most suitable donor image based on the consistency of chromaticity and complexity between the donor image and the acceptor image. Using all forged regions as candidate donors, we select the most suitable one for each CASIA v2.0 image and insert it at the most concealed position. For enlarging copy-move images in CASIA v2.0, we first find the authentic image of the forged region, then further find the corresponding copy-move image synthesized from this authentic image, and insert the forged region again at the most hidden position in this copy-move image. Each image of our enlarged CASIA v2.0 contains multiple forged regions, which may come from different images at the same time (possibly both from other images and from that image itself). These characteristics make the enlarged dataset more adaptable to complex forgery scenarios in practical applications. For removal operation, we randomly remove an annotated region from each ADE20k \cite{ref45} image and fill it using the SOTA inpainting method \cite{ref46}. We have generated $156006$ synthesized images ($140432$ for training, $7787$ for validation, and $7787$ for testing. Our dataset can be downloaded in \url{https://github.com/free1dom1/TBFormer}).}

\subsubsection{Testing data}
{We use four publicly available datasets, i.e., NIST16 \cite{ref47}, CASIA v1.0 \cite{ref44}, IMD20 \cite{ref48}, and Realistic \cite{ref49}, to evaluate the performance of our model. CASIA v1.0 contains splicing and copy-move images. NIST16, IMD20, and Realistic contain splicing, copy-move, and removal images.}

\subsubsection{Evaluation metrics}
{We use F1-score, IoU and AUC as evaluation metrics. $0.5$ is chosen as the threshold for all images when binarizing the predicted masks.}

\subsubsection{Implementation details}
{All the input images are resized to ${512\times 512}$. The feature extractor is initialized using the ViT model provided in \cite{ref50}, and the Transformer layers in the decoder are initialized using random weights from a truncated normal distribution. We use the SGD optimizer with the learning rate adjusted by the polynomial decay strategy ${lr=lr_{0}\left(1-iter_{current}/iter_{total}\right)^{0.9}}$, where ${iter_{current}}$ denotes the current number of iterations, ${iter_{total}}$ denotes the total number of iterations, and ${lr_{0}=0.001}$ denotes the initial learning rate. We set the batch size to $8$ and conduct $15$-epoch training, i.e., $263310$ iterations.}

\subsection{Ablation Study and Robustness Analysis}

\subsubsection{Ablation study}
\begin{table}[!htbp] 
\vspace{-1.0em}
\caption{Ablation study on Synthesized Dataset}
\vspace{-1.0em}
\centering
\begin{tabular}{ccccc}
\toprule
\multicolumn{1}{c}{Variants}&Precision&Recall&F1&IoU\\
\hline
\multicolumn{1}{c}{RGB-Only}&0.922&0.872&0.890&0.825\\
\multicolumn{1}{c}{RGB+Noise}&\pmb{0.924}&0.875&0.892&0.828\\
\multicolumn{1}{c}{RGB+Noise+AHFM}&0.917&\pmb{0.885}&\pmb{0.893}&\pmb{0.830}\\
\bottomrule
\end{tabular}
\label{tab1}
\vspace{-1.0em}
\end{table}
{In order to verify the effectiveness of the main modules, we set up different variants and conduct a series of experiments on the testing set of the synthesized dataset. Table \ref{tab1} reports the experimental results of different variants. ``RGB-Only'' indicates that only the features output by the last layer of the RGB branch are fed into the decoder, ``RGB+Noise'' means that a two-branch structure is used, but only the features output by the last layer of two branches are fed into the decoder after simply concatenation, and ``RGB+Noise+AHFM'' denotes the proposed method, i.e., TBFormer. The results can demonstrate that both the two-branch architecture and the AHFM module are helpful to improve the performance. The F1-score and IoU can be improved by adding each module. AHFM can improve the recall with precision sacrificing, which indicates that AHFM can reserve more information from multi-domain hierarchical features, while cause more inevitable false alarms.}

\subsubsection{Robustness analysis}
\begin{table}[!htbp] 
\vspace{-1.0em}
\caption{AUC scores of TBFormer on IMD20 under various distortions}
\vspace{-1.0em}
\centering
\begin{tabular}{ccc}
\toprule
\multicolumn{1}{c}{Distortion}&\multicolumn{2}{c}{AUC}\\
\hline
\multicolumn{1}{c}{no distortion}&\multicolumn{2}{c}{0.863}\\
\multicolumn{1}{c}{Resize(0.78×)}&0.855&-0.008\\
\multicolumn{1}{c}{Resize(0.25×)}&0.853&-0.010\\
\multicolumn{1}{c}{GaussianBlur(k=3)}&0.852&-0.011\\
\multicolumn{1}{c}{GaussianBlur(k=15)}&0.792&-0.071\\
\multicolumn{1}{c}{JPEGCompress(q=100)}&0.861&-0.002\\
\multicolumn{1}{c}{JPEGCompress(q=50)}&0.822&-0.041\\
\bottomrule
\end{tabular}
\label{tab2}
\vspace{-1.0em}
\end{table}
{We conduct various distortion transformations, e.g., resizing, JPEG compression and Gaussian blur on the IMD20 dataset to evaluate the robustness of the model, and the experimental results are shown in Table \ref{tab2}. From Table \ref{tab2}, we can see that the AUC scores do not significantly decrease under different distortions, which can demonstrate the robustness of our TBFormer.}

\subsection{Comparison With State-of-the-art Methods}

\begin{table}[!htbp] 
\vspace{-1.0em}
\caption{Comparison With State-of-the-art Methods}
\vspace{-1.0em}
\centering
\begin{tabular}{ccccccc}
\toprule
\multicolumn{1}{c}{\multirow{2}{*}{Method}}&\multicolumn{2}{c}{NIST16}&\multicolumn{2}{c}{CASIA v1.0}&IMD20&Realistic\\
\multicolumn{1}{c}{}&AUC&F1&AUC&F1&AUC&AUC\\
\hline
\multicolumn{1}{c}{RGB-N}&0.937&0.722&0.795&0.408&-&-\\
\multicolumn{1}{c}{ManTra-Net}&-&-&-&-&0.748&0.680\\
\multicolumn{1}{c}{SPAN}&0.961&0.582&0.838&0.382&0.750&-\\
\multicolumn{1}{c}{MVSS-Net}&-&-&0.887&0.539&0.814&0.641\\
\multicolumn{1}{c}{PSCC-Net}&0.996&0.819&0.875&0.554&0.806&0.542\\
\multicolumn{1}{c}{ObjectFormer}&0.996&0.824&0.882&0.579&0.821&-\\
\multicolumn{1}{c}{TBFormer}&\pmb{0.997}&\pmb{0.834}&\pmb{0.955}&\pmb{0.696}&\pmb{0.863}&\pmb{0.738}\\
\bottomrule
\end{tabular}
\label{tab3}
\vspace{-1.0em}
\end{table}

\begin{figure}
\centerline{\includegraphics[width=\columnwidth]{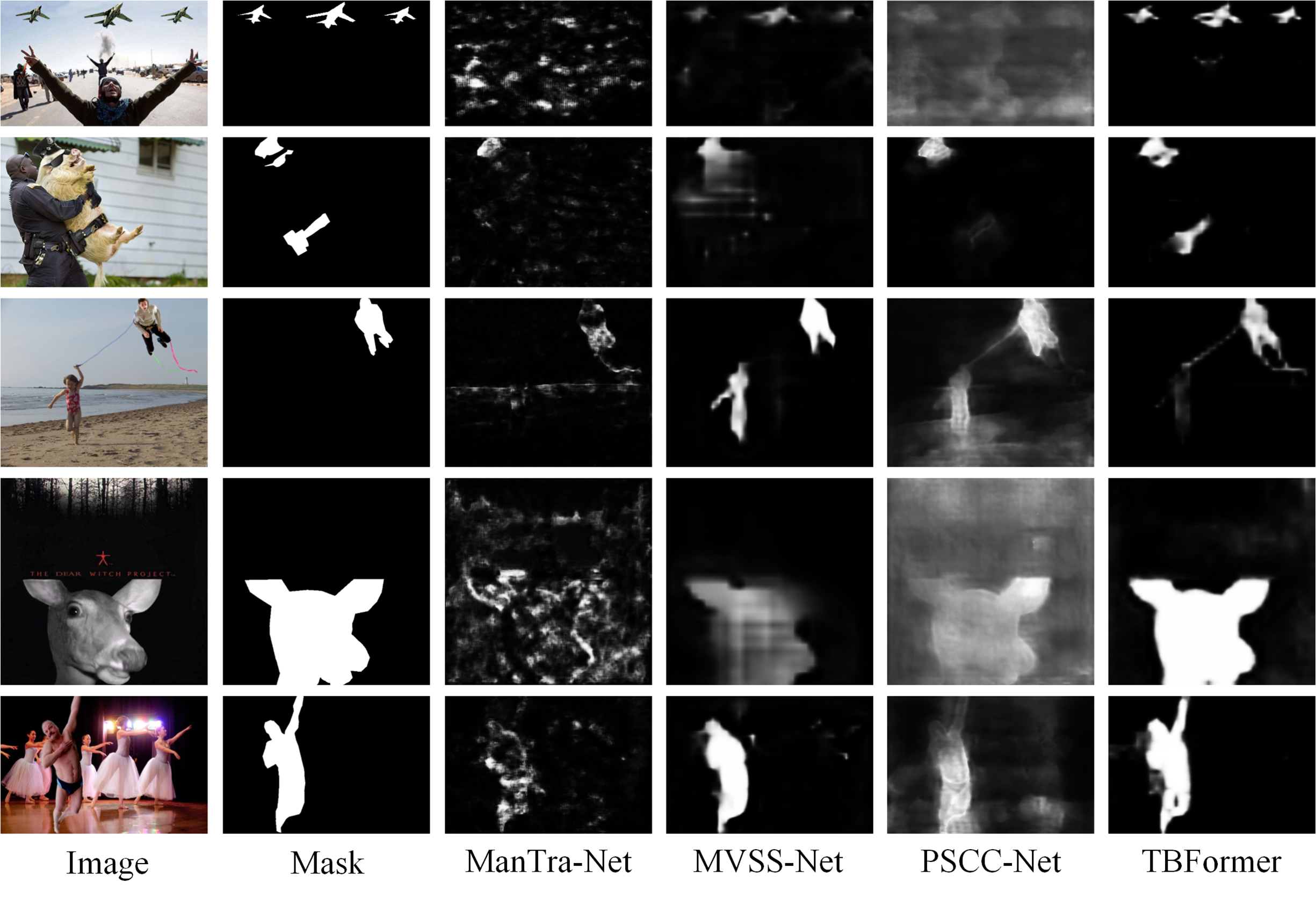}}
\vspace{-1.0em}
\caption{Visual comparisons with the state-of-the-art methods.}
\label{fig3}
\vspace{-1.0em}
\end{figure}

TBFormer is compared with six state-of-the-art methods, i.e., RGB-N \cite{ref26}, ManTra-Net \cite{ref25}, SPAN \cite{ref15}, MVSS-Net \cite{ref21}, PSCC-Net \cite{ref20}, and ObjectFormer \cite{ref35}. Table \ref{tab3} reports the compared results on four publicly available datasets, and Fig. \ref{fig3} visualizes predicted masks of the methods with publicly available codes. On the NIST16 dataset, RGB-N, SPAN, PSCC-Net, and ObjectFormer are fine-tuned, and we follow the same training/testing splits for fine-tuning the model to make a fair comparison. On the CASIA dataset, RGB-N, SPAN, PSCC-Net, and ObjectFormer are fine-tuned on CASIA v2.0 and tested on CASIA v1.0. The training dataset of MVSS-Net and our synthesized dataset are generated from CASIA v2.0, so the results of MVSS-Net and TBFormer on CASIA v1.0 are generated by the models without fine-tuning. The results of MVSS-Net on all datasets, the scores of ManTra-Net and SPAN on the IMD20 dataset, and the results of compared methods on the Realistic dataset are obtained by the pre-trained models released by the authors, and the rest of scores are borrowed from their original papers. Table \ref{tab3} shows that TBFormer achieves the best performance on each dataset, and it also can be seen in Fig. \ref{fig3} that TBFormer can locate forged regions more accurately.

\section{CONCLUSION}

In this paper, we introduce a novel Transformer-based image forgery localization model, named as TBFormer, which can achieve superior performance. TBFormer uses two Transformer branches to extract RGB and noise features independently to fully explore the potential forgery cues. The Attention-aware Hierarchical-feature Fusion Module (AHFM) is proposed for effectively integrating hierarchical features extracted from RGB and noise domains. Finally, the predicted mask is reconstructed by the Transformer decoder. In the future, TBFormer can be further improved by considering edge artifacts or other potential forgery cues.

\end{document}